\documentclass[journal]{IEEEtran}
\usepackage{lineno,hyperref}
\usepackage{bbm}
\usepackage{bm}
\usepackage{amsfonts}
\usepackage{amsmath}
\usepackage{mathrsfs}
\usepackage{amssymb}
\usepackage{enumerate}
\usepackage{graphicx}
\usepackage{subfigure}
\usepackage{booktabs}
\usepackage{graphicx}
\usepackage{setspace}
\usepackage{algorithm}
\usepackage{algorithmic}
\usepackage{setspace}
\usepackage{multirow}
\usepackage{color}
\usepackage{cite}
\usepackage{array}
\usepackage{times}
\usepackage{mathptmx}
\usepackage{dsfont}
\ifCLASSINFOpdf
\else
\fi
\begin{document}
\title{Domain-adaptive Person Re-identification without Cross-camera Paired Samples}
\author{Huafeng~Li$^{\scriptscriptstyle \ast}$, Yanmei~Mao$^{\scriptscriptstyle \ast}$, Yafei Zhang, Guanqiu~Qi, and Zhengtao~Yu \IEEEmembership{}
\thanks{This work was supported in part by the National Natural Science Foundation of China under Grant 61966021, Grant 61772455, Grant 61562053,
Grant 61572486, Grant U1713213, and Grant 61501177, in part by the National Key Research and Development Plan Project under Grant
2018YFC0830105 and Grant 2018YFC0830100, in part by Yunnan Natural Science Funds under Grant 2018FY001(-013), in part by the Program for Excellent Young Talents of National Natural Science Foundation of Yunnan University under Grant 2018YDJQ004, in part by the Yunnan Natural Science Funds under Grant 2017FB094,
and in part by the Program for Excellent Young Talents of Yunnan University under Grant WX069051.}
\thanks{H. Li, Y. Mao, Y. Zhang, and Z. Yu are with the Faculty of Information Engineering and Automation, Kunming University of Science and Technology, Kunming 650500, China. (E-mail:lhfchina99@kust.edu.cn (H. Li); maoyanmeicy@163.com (Y. Mao); zyfeimail@163.com (Y. Zhang); ztyu@hotmail.com (Z. Yu))}
\thanks{G. Qi is affiliated with Computer Information Systems Department, State University of New York at Buffalo State, Buffalo, NY 14222, USA. (E-mail:qig@buffalostate.edu)}
\thanks{}
\thanks{ ${\ast}$ indicates contributed to this work equally.}
\thanks{Manuscript received xxxx;}}
\markboth{Journal of \LaTeX\ Class Files}%
{Shell \MakeLowercase{\textit{et al.}}}
\maketitle
\begin{abstract}
Existing person re-identification (re-ID) research mainly focuses on pedestrian identity matching across cameras in adjacent areas. However, in reality, it is inevitable to face the problem of pedestrian identity matching across long-distance scenes.
The cross-camera pedestrian samples collected from long-distance scenes often have no positive samples. It is extremely challenging to use cross-camera negative samples to achieve cross-region pedestrian identity matching.
Therefore, a novel domain-adaptive person re-ID method that focuses on cross-camera consistent discriminative feature learning under the supervision of unpaired samples is proposed. This method mainly includes category synergy co-promotion module (CSCM) and cross-camera consistent feature learning module (CCFLM). In CSCM, a task-specific feature recombination (FRT) mechanism is proposed. This mechanism first groups features according to their contributions to specific tasks. Then an interactive promotion learning (IPL) scheme between feature groups is developed and embedded in this mechanism to enhance feature discriminability. Since the control parameters of the specific task model are reduced after division by task, the generalization ability of the model is improved. 
In CCFLM, instance-level feature distribution alignment and cross-camera identity consistent learning methods are constructed. 
Therefore, the supervised model training is achieved under the style supervision of the target domain by exchanging styles between source-domain samples and target-domain samples, and the challenges caused by the lack of cross-camera paired samples are solved by utilizing cross-camera similar samples. In experiments, three challenging datasets are used as target domains, and the effectiveness of the proposed method is demonstrated through four experimental settings.
\end{abstract}
\IEEEpeerreviewmaketitle
\section{Introduction}
Person re-ID is a technique used to determine whether pedestrians under non-overlapping cameras have the same identity. Due to its wide application prospect in criminal suspect tracking and missing person search, person re-ID has attracted significant attention from researchers, and a large number of effective methods have been proposed. Among these methods, supervised person re-ID was first proposed\cite{43,44,45,46,47,48,49,50,51,52}. Under the supervision of a large-scale labeled training samples, the recognition performance quickly reaches a certain level with the continuous advancement of deep learning technology. The domain adaptability of the model obtained by this supervised training is poor, and direct deployment to new datasets always results in performance degradation due to the domain shift between source domain and target domain\cite{53,54,55}. As the most direct and effective way to solve this problem, large-scale training samples are labeled on the target domain, and the recognition model is supervised for training. However, manually labeling large-scale training samples is extremely time-consuming and laborious\cite{31}. Therefore, fully unsupervised and domain-adaptive person re-ID is proposed.
\begin{figure}[t!]
\centering
\includegraphics[width=3.4in,height=1.9in]{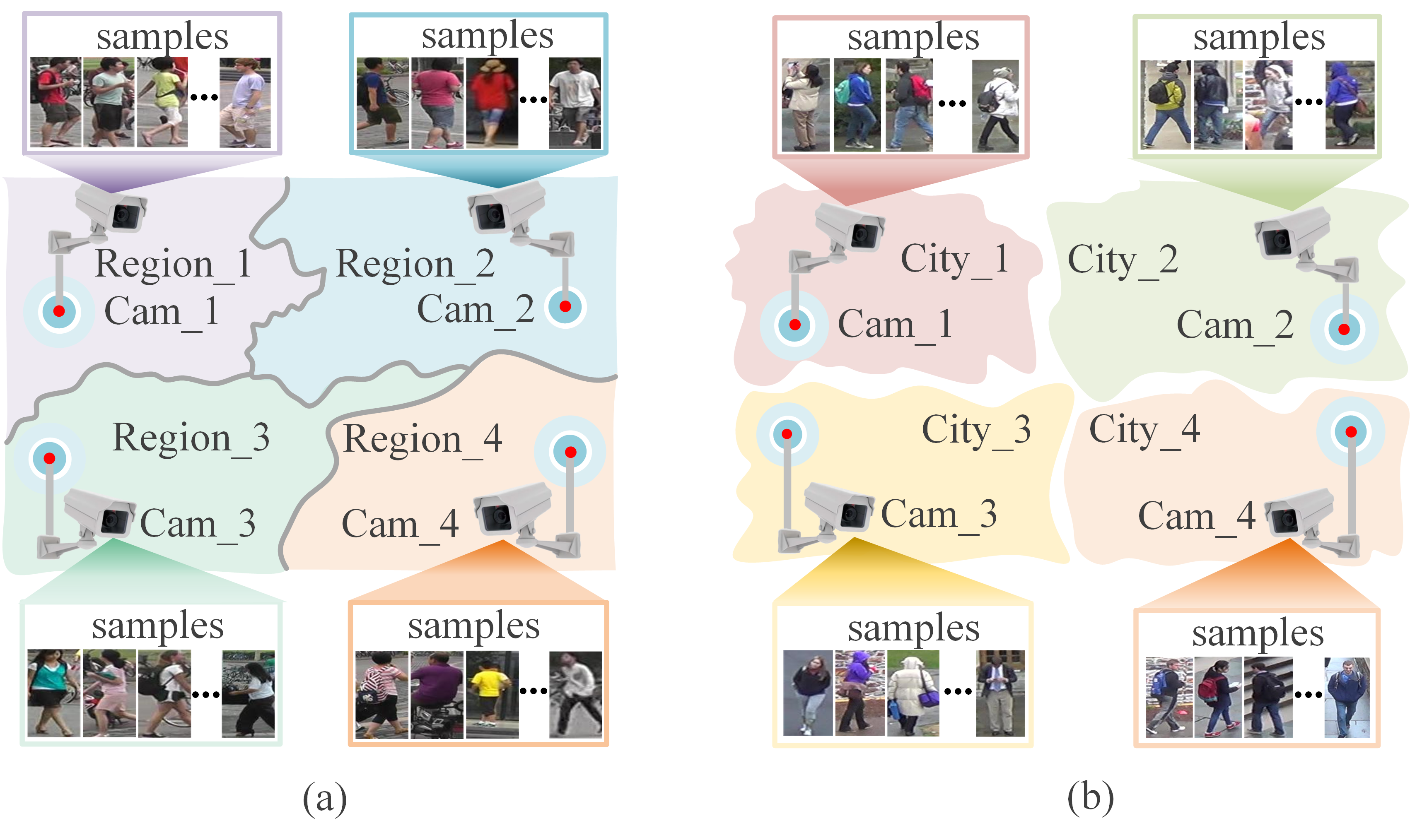}
\caption{Illustration of the pedestrian samples captured by different cameras in long-distance scenes. Pedestrians with the same identity across cameras do not appear in long-distance scenes. (a) Cross-region scenes. (b) Cross-city scenes.}
\label{label}
\end{figure}

\begin{figure*}[t!]
\centering
\includegraphics[width=6.7in,height=3.7in]{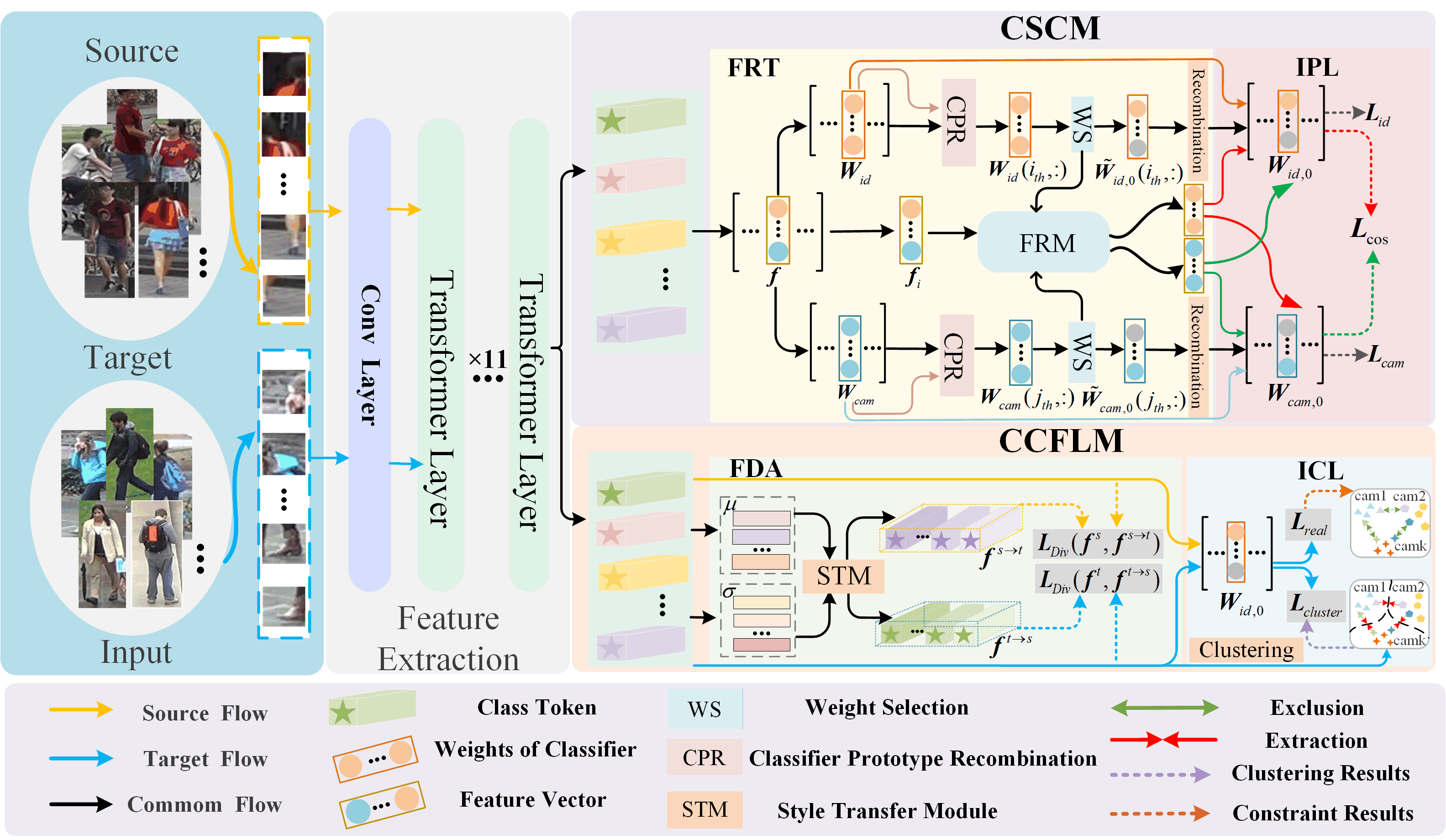}
\caption{The overview of the proposed method. In the model training process, the labeled source-domain samples and cross-camera unpaired target-domain samples are first sent to the feature extraction network $\bm E$ for pre-training. So, the pre-trained $\bm E$ has certain pedestrian and camera recognition capabilities. After that, the features extracted by the pre-trained $\bm E$ are grouped, and the interactive promotion learning between grouped features improves feature discriminability. Additionally, the features extracted by $\bm E$ are sent to CCFLM to guarantee distribution consistency and identity consistency among cross-camera features.}
\label{labe2}
\end{figure*}

In general, fully unsupervised methods directly predict pseudo-labels for the training set of the target dataset and utilize the predicted pseudo-labels to supervise model training. Such methods require high reliability of pseudo-labels. If there is a lot of noise in the pseudo-labels, it causes a significant drop in model performance on the target dataset. Domain-adaptive methods mainly use data from source and target domains to train re-ID models. During this process, the source-domain data is labeled while the target-domain data is unlabeled. Source-domain and target-domain data supervise model training by annotated true labels and predicted pseudo-labels, respectively.
Therefore, domain adaptative methods are theoretically more stable than fully unsupervised methods, thus attracting the attention of researchers.

Existing domain-adaptive person re-ID methods mainly address short-distance cross-camera pedestrian matching.
This has a positive impact on reducing noisy labels, as pedestrians with the same identity are more likely to appear in short-distance
cross-camera views. However, it actually needs to match pedestrian identities across long-distance scenes, as shown in Fig. \ref{label}. If pedestrian A appears in the region-1 of a city, which region of the city will he appear in again? This involves the problem of long-distance cross-camera pedestrian identity matching. In such scenes, it is highly likely that cameras across long-distance scenes do not capture pedestrians with the same identity. If existing domain-adaptive person re-ID methods are directly deployed to the long-distance scenes, the original performance drops significantly,
because the predicted pseudo-labels are all noisy labels. In view of the lack of cross-camera paired samples, some researchers proposed to predict features from one camera view to another
and used the prediction results and original features to form cross-camera sample pairs for supervised model training. However, these methods rely heavily on the quality of prediction results for missing samples across cameras.

To address the above issues, this paper proposes a domain-adaptive person re-ID method without cross-camera paired samples. Unlike existing methods, the proposed method does not need to use pedestrians from a single camera view to generate cross-camera samples or features. It addresses the challenge of missing cross-camera pedestrian sample pairs by improving the domain adaptability of the model on cross-region datasets. Specifically, FRT mechanism is proposed to learn cross-camera discriminative features. This mechanism divides features into two categories according to their contributions in pedestrian identity classification, pedestrian identity-related features and camera identity-related features. This design reduces the features that contribute less to pedestrian identity classification, thereby promoting the role of features that contribute more to pedestrian identity classification. In addition, the control parameters of the model are reduced, which further improves the generalization ability of the model.
On this basis, IPL scheme is designed to realize mutual promotion learning between features. This mechanism can not only promote the improvement of feature discriminability but also achieve the separation of camera and pedestrian identity information, reducing the interference of camera information on pedestrian identity-related features.

To further address the challenge caused by the lack of cross-camera positive samples, cross-camera consistent feature learning is implemented within the framework. By exchanging the style of source and target domains, the alignment of instance-level data distribution is achieved, and the supervised model training under the style of the target domain improves the domain adaptability of the model. To enable the model to extract the same pedestrian features across cameras, a cluster-based consistent feature learning method is proposed. This method regards pedestrians with consistent identities across cameras as similar pedestrians and compensates for the impact of missing cross-camera positive samples on model training by reducing the difference of intra-class features. Additionally, identity-inconsistent features are further excluded based on the prior knowledge that pedestrians with the same identity do not exist across long-distance scenes. This ensures that the model can still extract the features of the same identity and exclude the features of different identities without cross-camera positive samples. In summary, this paper has three main contributions as follows.
\begin{itemize}
\item A task-specific feature recombination mechanism is proposed.  According to the contributions to pedestrian identity classification based on features, the mechanism divides features into two categories related to pedestrian identity and camera identity, and performs interactive promotion learning between the two categories. This not only improves the discriminability of the features, but also improves the generalization of the model.

\item Aiming at the problem of missing cross-camera paired samples, an instance-level feature distribution alignment and cross-camera identity consistent feature learning method is proposed. Exchanging source-domain and target-domain camera styles and constraining the category consistency of similar pedestrians across cameras are applied to address the inability to train models supervised with target-domain data.

\item The experimental results under four experimental settings on three datasets, i.e., Market-SCT \cite{1}, DukeMTMC-SCT (Duke-SCT) \cite{1}, and MSMT17-SCT (MSMT-SCT) \cite{2}, confirm that the proposed method achieves superior recognition performance compared with existing methods on pedestrian identity matching in long-distance scenes.
\end{itemize}

The rest of this paper is organized as follows. Section \uppercase\expandafter{\romannumeral2} discusses related work; Section \uppercase\expandafter{\romannumeral3} specifies the proposed method; Section \uppercase\expandafter{\romannumeral4} shows the experimental results and correspoding analysis; and Section \uppercase\expandafter{\romannumeral5} concludes this paper.
\section{Related Work}
\subsection{Intra-Camera Supervised Person Re-ID}
Due to the strong application prospects, person re-ID across long-distance scenes has received significant attention from researchers in the past two years. In 2020, Zhang et al. \cite{1} first proposed the application requirements of the problems across long-distance scenes and designed a single-camera training method to solve the problem caused by the lack of cross-camera paired samples. This method guarantees the discriminability of features by extracting hard-positive samples in the camera and excluding hard-negative sample pairs within-camera and cross-cameras. In cross-camera person identity matching, mitigating cross-camera intra-class variations is significant for person re-ID. However, this method ignores the impact of the above-mentioned problem on recognition performance. Therefore, Ge et al.  \cite{2} proposed a method to predict cross-camera pedestrian features of single-camera pedestrians. The predicted results are combined with the original features to form cross-camera positive sample pairs and used to supervise the model training. This method relies heavily on the predicted cross-camera pedestrian features. If the predictions are not good, it most likely hurts the model performance. To alleviate this problem, Wu et al. \cite{3} proposed a camera-conditioned stable feature generation method, which reduces the impact of unideal features on model performance by improving the quality of generated features. Although this method improves the quality of generated cross-camera sample, the problem of model performance being limited by the quality of generated samples remains unresolved. Different from the above methods, the proposed method belongs to the domain adaptive method. It addresses the challenge of missing cross-camera paired samples by improving model generalization and domain adaptability.
\subsection{Domain-Adaptive Person Re-ID}
The domain-adaptive person re-ID method trains the model through labeled source-domain samples and unlabeled target-domain samples, so that the model has good performance in the target domain. Since this method does not need to label samples in the target domain, which significantly reduces the labor cost of labeling training samples, it has attracted the attention of researchers and proposed a series of effective methods. Existing methods can be roughly classified into three categories, methods based on external model assistance, methods based on domain-adaptive feature extraction, and methods based on pseudo-label prediction. Methods based on external model assistance usually use an external model to assist re-ID models to gain domain adaptability. Commonly used external models include CamStyle transfer model \cite{4}, pedestrian pose extraction model \cite{5}, etc. In particular, methods such as PT-GAN \cite{6}, SPGAN \cite{7}, ATNet \cite{8}, and CR-GAN \cite{9} all use the CamStyle transfer model to transfer the style of training samples from the source domain to the target domain, so that the model can be trained in a supervised manner on the transferred samples. However, the samples generated after style transfer are prone to corrupt information related to pedestrian identity cues. Utilizing such samples to train a model can easily degrade recognition performance. In pedestrian pose-assisted methods, the pose estimation model is usually used to estimate the pose information of pedestrians in the training set and combined with the generative adversarial mechanism to achieve pedestrian pose-invariant and domain-aligned feature extraction. In addition, to obtain discriminative features with consistent distribution, Yang et al. \cite{10} proposed the part-aware progressive unsupervised domain adaptation person re-ID method. This method utilizes a pedestrian's pose estimation model to perceive pedestrian body parts. These methods, which rely on external models for data preprocessing, are not conducive to the deployment in real-world scenes.

Methods based on domain-adaptive feature extraction often address the impact of domain discrepancy on recognition performance by learning transferable features. In particular, Yang et al. \cite{11} proposed a patch-based unsupervised learning framework for obtaining transferable discriminative features. This method exploits the similarity between patches to learn features that are transferable across cameras. Wu et al. \cite{12} proposed a camera-aware similarity consistency loss to learn consistent pairwise similarity distributions to address domain shift. Qi et al. \cite{13} proposed an unsupervised camera-aware domain adaptation framework to address domain shift between source and target domains. Zou et al. \cite{14} proposed to purify the target domain representation space through feature disentanglement, thereby enhancing the domain adaptability of the model. Chen et al.\cite{15} used the domain invariance of pedestrian attribute information to improve the performance of visual features in domain-adaptive pedestrian matching. Dong et al. \cite{16} proposed a triple adversarial learning and imaginative reasoning mechanism to obtain domain-adaptive discriminative features. Li et al.\cite{17} proposed a mutual-promotion learning method for feature disentanglement, which achieved the separation of domain-invariant features and domain-related features, improving the performance of the model on the target domain. Although methods based on domain-adaptive feature extraction are effective, their performance is limited due to the underutilization of unlabeled samples from the target domain in model training.

To make full use of unlabeled samples from the target domain in model training, methods based on pseudo-label prediction have been proposed. In particular, Ge et al. \cite{18} proposed an unsupervised mutual mean-teaching (MMT) framework for domain-adaptive person re-ID. The framework improves the accuracy of label prediction through offline refinement of hard pseudo-labels and online refinement of soft pseudo-labels using an alternating training scheme. Yang et al. \cite{19} designed an asymmetric co-teaching framework. The framework suppresses noisy labels through the collaboration of two models. Zhai et al. \cite{20} proposed an augmented discriminative clustering (AD-Cluster) technique to predict pseudo-labels of pedestrians in the target domain. Zeng et al. \cite{21} proposed a hierarchical clustering method with hard-batch triplet loss to improve the quality of pseudo-label prediction. For the problem of model non-convergence in cluster-based pseudo-label prediction methods. Ji et al. \cite{22} proposed an attention-driven two-stage clustering method. To suppress the noise in pseudo-labels, Zhao et al. \cite{23} proposed a noise resistible mutual training (NRMT) mechanism to predict pseudo-labels through interactive instance selection. Shi et al. \cite{24} proposed a reliability exploration with self-ensemble learning (RESL) framework to improve the reliability of pseudo-label prediction in domain-adaptive person re-ID.

To improve the adaptation of the model on the target domain, Chen et al. \cite{25} combined data augmentation methods with a multi-label assignment strategy to decouple semantic features from the source domain. Additionally, a pre-trained model was used to extract various semantic features from the target dataset, and each semantic feature is regarded as a specific domain. The features of each domain are then clustered, and the connections between different clusters are used for self-distillation to generate more reliable pseudo-labels. Li et al. \cite{26} exploited a logical reasoning mechanism that utilizes the logical relationships between samples to refine cross-camera pseudo-labels. When the training set in the target dataset contains a large number of cross-camera positive samples, the above-mentioned methods can predict pseudo-labels relatively reliably. However, across long-distance scenes, the possibility of the same pedestrian appearing under two cameras is extremely low, resulting in the target-domain samples participating in model training being negative samples across cameras.
\section{The Proposed Method}
\subsection{Overview of the Proposed Framework}
As shown in Fig. \ref{labe2}, the proposed method consists of the feature extraction network $\bm E$, category synergy co-promotion module (CSCM), and cross-camera consistent feature learning module (CCFLM). The $\bm E$ is used to extract features from the source-domain and target-domain samples. The CSCM is composed of feature recombination by task (FRT) and interactive promotion learning (IPL). According to the contributions of features to pedestrian and camera identification, FRT divides features into two groups. One group is related to pedestrian identity and the other group is related to camera identity. Then the IPL of features is developed and carried out between the two groups to improve the discriminability of the features. The CCFLM includes the feature distribution alignment (FDA) and cross-camera identity consistent feature learning (ICL) under the supervision of cross-camera unpaired samples. The FDA is mainly used to make the cross-domain features share the same probability distribution. The ICL is mainly used to make the features learned by the model under the supervision of unpaired samples have cross-camera identity consistency.
\subsection{Feature Extraction Network and Pre-training}
Assume that the source-domain data sample set is $ \bm D_{s}=\{\bm x_{s, i}, y_{s, i}, c_{s, i}\}_{i=1}^{N_{s}}$, where $\bm x_{s, i}$ represents the $i$-th sample from $\bm D_{s}$, $y_{s, i}\in\{1, 2, \cdots, n_{s}\}$  and $ c_{s, i}\in\{1, 2, \cdots, k_{s}\}$ represent the identity label and camera label of the sample $\bm x_{s, i}$, $N_{s}$ is the total number of samples in $\bm D_{s}$, $n_{s}$ denotes the total category of pedestrian identity, and and $k_{s}$ denotes the total number of cameras. Similarly, the training set in the target domain is defined as $\bm D_{t}=\{\bm x_{t, j}, y_{t, j}, c_{t, j}\}_{j=1}^{N_{t}}$, where $y_{t, j}\in\{1, 2, \cdots, n_{t}\}$  and $c_{t, j}\in\{1, 2, \cdots, k_{t}\}$. In cross-region person re-ID, according to the protocol of \cite{1}, it assumes that the training set or domain adaptation in $\bm D_{t}$ has no paired images across camera views, i.e., for $\forall i, j$, $y_{t, i}\neq y_{t, j}$ when $c_{t, i}\neq c_{t, j}$. Images of pedestrians with the same identity only appear in a single camera, i.e., $\exists i, j$, such that $y_{t, i}=y_{t, j}$ when $c_{t, i}=c_{t, j}$.

In the feature extraction of pedestrian images, the TransReID \cite{56} framework is used as the backbone denoted as $\bm E$. The network consists of 12 transformer layers. In the proposed method, it is necessary to analyze the contribution of extracted features to specific tasks and group them accordingly. Therefore, it is necessary to make $\bm E$ have the initial recognition ability. Let $\bm E(\bm x_{s,i})$ be the global token of the $i$-th pedestrian image $\bm x_{s,i}\in {\bm D_s}$, and $\bm E(\bm x_{s,i}^k)$ be the $k$-th local token. The feature extraction network $\bm E$ can realize pre-training by minimizing Eq.(1).
\begin{equation}
\begin{aligned}
	\bm L_{id_{1}}(\bm E)= \frac{1}{{{n_s}}}\sum\limits_{i = 1}^{{n_s}} {Ce\bigg({{\bm{W}}_{id}}\Big({\bm{E}}\big({\bm x_{s,i}}\big),{{{y}}_{s,i}}\Big)\bigg) + Tri\Big({\bm{E}}\big(\bm x_{s,i}^{}\big)\Big)}\\
	+\frac{1}{{{n_s}K}}\sum\limits_{i = 1}^{{n_s}} {\sum\limits_{k = 1}^K {Ce\bigg({{\bm{W}}_{id}}\Big({\bm{E}}\big(\bm x_{s,i}^k\big),{{{y}}_{s,i}}\Big)\bigg) + Tri\Big({\bm{E}}\big(\bm x_{s,i}^k\big)\Big)}}
\end{aligned},
\end{equation}
where $Ce(\cdot)$ and $Tri(\cdot)$ are the cross-entropy loss and triplet loss functions respectively, ${{\bm{W}}_{id}}\in \mathbb{R}^{m \times n}$ is the identity classifier, $m$ is the total number of pedestrian identities, and $n$ is the number of channels of the feature vector. ${y}_{s,i}$ represents the identity label of the $i$-th sample from the source domain, and ${n_s}$ is the number of source-domain samples in a mini-batch.

Since the pedestrian identity classifier and the camera classifier in the proposed method need to promote each other to realize the separation of pedestrian identity and camera information, it is also necessary to pre-train the camera classifier.
\begin{equation}
\begin{aligned}
	\bm L_{cam}({\bm W}_{cam})= \sum\limits_{i = 1}^{{n_s} + {n_t}} Ce\bigg({\bm W}_{cam}\Big({\bm E}\big(\bm x_i^{}\big),{{c}_i}\Big)\bigg)
\end{aligned},
\end{equation}
where $\bm x_i^{} \in {\bm D_s} \cup {\bm D_t}$, ${c_i}$ is the camera label of ${\bm x_i}$, ${{\bm{W}}_{cam}}$ is the camera classifier, and ${n_t}$ is the number of target-domain samples in a mini-batch.

\subsection{Category Synergy Co-promotion Module}
In the domain-adaptive person re-ID, pedestrian discriminative feature extraction, which is not affected by domain shift, plays a crucial role in improving the domain generalization ability of the model. In multi-classification tasks, different features contribute differently to a  specific classification task. If these features can be grouped by their contributions to a specific classification task, it is conductive to enhancing the ability of the network to extract task-specific discriminative features. Therefore, this paper proposes a CSCM. This module is mainly composed of FRT and IPL.
\subsubsection{Task-specific Feature Recombination}
\begin{figure}[t!]
\centering
\includegraphics[width=3.0in,height=2.0in]{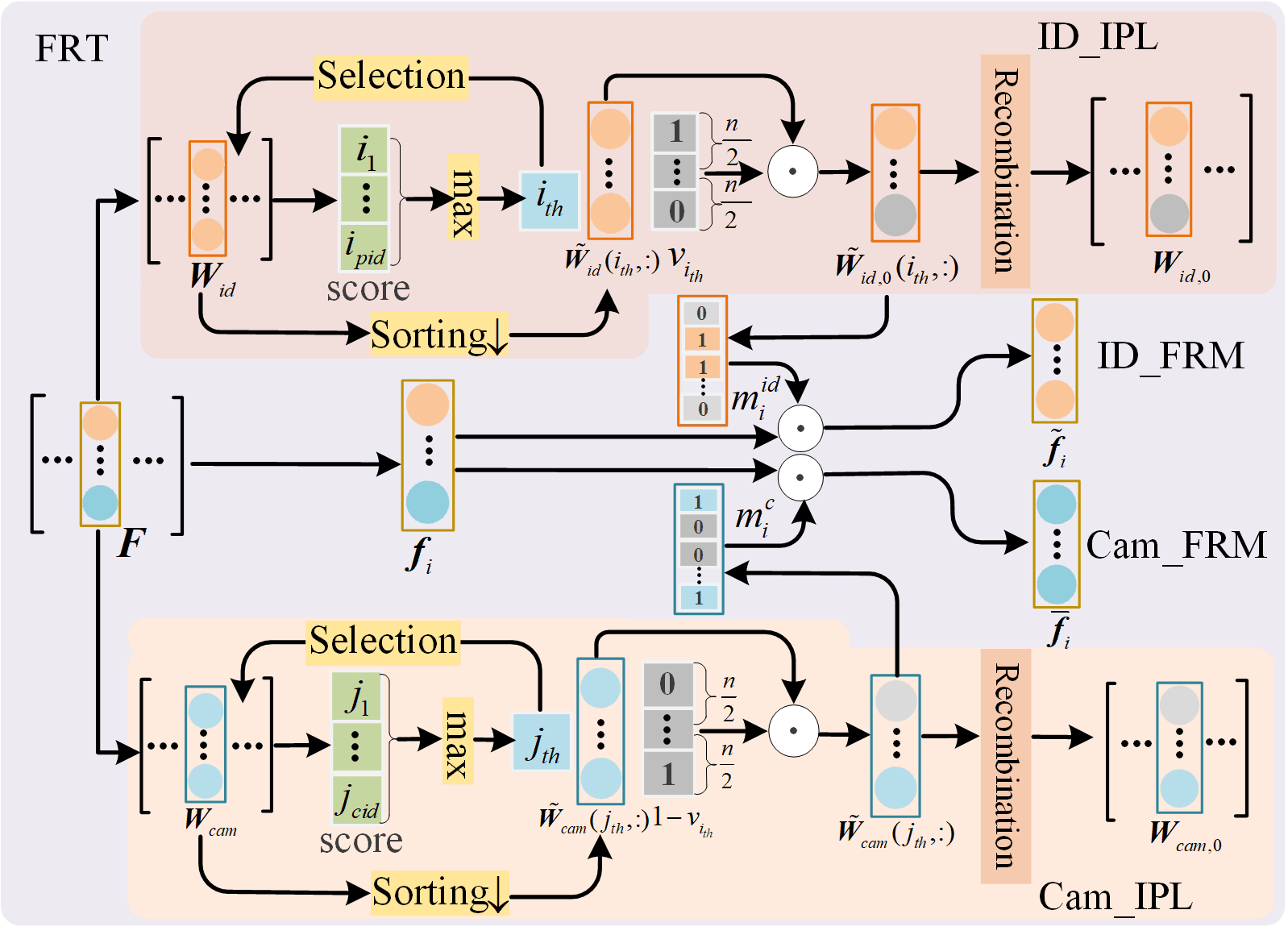}
\caption{Illustration of task-specific feature recombination.}
\label{label3}
\end{figure}

Fig. \ref{label3} shows the specific process of FRT. It consists of the feature recombination module by pedestrian identity (ID-FRM) and feature recombination module by camera identity (Cam-FRM). The former is mainly used to improve the discriminability of pedestrian identity-related features, and the latter is mainly used to assist the former in excluding camera-related domain information in features. Assume that the output feature of the source-domain sample $\bm x_{s,i}$ passing through the feature extraction network $\bm E$ is $\bm f_{s,i}$. The classification result obtained after feeding $\bm f_{s,i}$ into the pre-trained pedestrian classifier $\bm W_{id}$ is given as follows.
\begin{equation}
\begin{aligned}
	\bm [{p_1},{p_2}, \cdots ,{p_m}]^T = \bm W_{id}(\bm f_{s,i})
\end{aligned},
\end{equation}
where $[{p_1},{p_2}, \cdots ,{p_m}]$ is the probability that $\bm x_{s,i}$ belongs to each identity of the source-domain pedestrian. Let $i_{th}$ be the position index of the largest value among $[{p_1},{p_2}, \cdots ,{p_m}]$ in $[{p_1},{p_2}, \cdots ,{p_m}]^T$ as follows.
\begin{equation}
\begin{aligned}
	  i_{th} = \arg \max \{ {p_1},{p_2}, \cdots ,{p_m}\}
\end{aligned},
\end{equation}

After obtaining ${i_{th}}$, the features are extracted from $\bm W_{id}$ that make significant contributions to the correct classification of $\bm {x_{s,i}}$, while excluding the remaining features. In the subsequent training of $\bm E$, the extracted features are encouraged to play a greater role in person identification, so as to implement network augmentation in discriminative feature extraction. Therefore, this paper looks for the vector $\bm W_{id}({i_{th}},:)$ corresponding to the ${i_{th}}$ identity from the identity classifier $\bm W_{id}$. For the convenience of subsequent operations, $\bm W_{id}({i_{th}},:)$ is marked as follows.
 \begin{equation}
\begin{aligned}
	\bm W_{id}(i_{th},:) = [{w_{{i_{th}},1}},{w_{{i_{th}},2}}, \cdots ,{w_{{i_{th}},n}}]
\end{aligned},
\end{equation}

To find the top $k$ features that contribute significantly to the correct classification, the values in ${\bm{W}}_{id}^{}({i_{th}},:)$ are arranged in order of size as follows.
\begin{equation}
\begin{aligned}
	{\bm{\tilde W}}_{id}^{}({i_{th}},:)
	&= {\mathop{\rm sort}\nolimits}  \downarrow ([{ w_{{i_{th}},1}},{ w_{{i_{th}},2}}, \cdots ,{ w_{{i_{th}},n}}])\\
	&= [{{\tilde w}_{{i_{th}},1}},{{\tilde w}_{{i_{th}},2}}, \cdots ,{{\tilde w}_{{i_{th}},n}}]
\end{aligned},
\end{equation}
where $\rm{sort}\downarrow$ represents the descending order operation.

To emphasize the role of features that contribute more to identity classification, this paper proposes to perform link inactivation on features that contribute less to identity classification. Let $\bm{v}_{i_{th}} = {\rm{[}}\underbrace {{\rm{1,1,}} \cdots 1}_{\frac{n}{2}}{\rm{,}}\underbrace {{\rm{0,0,}} \cdots 0}_{\frac{n}{2}}{\rm{]}}$. ${\bm{\tilde W}}_{id}^{}({i_{th}},:)$ after inactivation can be expressed as follows.
\begin{equation}
\begin{aligned}
	{\bm{\tilde W}}_{id,0}^{}({i_{th}},:) = {\bm{\tilde W}}_{id}^{}({i_{th}},:) \odot {{\bm{v}}_{{i_{th}}}}
\end{aligned},
\end{equation}
where $\odot$ is element-wise dot product.

The value in ${\bm{\tilde W}}_{id,0}^{}({i_{th}},:)$ is put back to the original position of the value in ${\bm{W}}_{id}^{}({i_{th}},:)$.
The classifier ${\bm{W}}_{id,0}^{}({i_{th}},:)$ is obtained.
When ${i_{th}}$ traverses all pedestrian identities in the source domain and the target domain, the obtained classifier is denoted as ${\bm{W}}_{id,0}^{}$.

To realize the feature discriminability improvement of the corresponding pedestrian image camera category, this paper proposes the following method to construct the deactivated camera classifier ${\bm{W}}_{cam,0}^{}$.
\begin{equation}
    \bm W_{cam,0}^{}(i,j)= \begin{cases}
	 \bm W_{cam,0}^{}(i,j), & \text {if $\bm W_{id,0}^{}(i,j) = 0$} \\
	$0$, &\text{otherwise}
\end{cases},
\end{equation}

The $\bm W_{cam,0}^{}$ constructed in this way can effectively prevent $\bm W_{cam,0}^{}$ and $\bm W_{id,0}^{}$ from using the same features to classify camera and pedestrian identities. This also enables the separation of pedestrian identity information from camera identity information within a feature extraction network framework. To remove camera information from pedestrian identity-related features and improve the domain adaptability of the model, the feature $\bm f_{s,i}$ is re-divided into the feature $\bm{\tilde f}_{s,i} \in \mathbb{R}^{1 \times n/2}$ for pedestrian identification and the feature $\bm{\bar f}_{s,i} \in \mathbb{R}^{1 \times n/2}$ for camera identification.

First, the features of $\bm f_{s,i}$ are grouped  as follows.
\begin{equation}
\begin{aligned}
   {{\bm{\tilde f'}}_{s,i}} = {{\bm f}_{s,i}} \odot {\bm m}_{s,i}^{id}
\end{aligned},
\end{equation}
\begin{equation}
\begin{aligned}
    {{\bm{\bar f'}}_{s,i}} = {{\bm f}_{s,i}} \odot {\bm m}_{s,i}^c
\end{aligned},
\end{equation}
where ${\bm{m}}_{s,i}^{id}({\bm{m}}_{s,i}^c)$ is a vector composed of 0 and 1. 

When the element value at a certain position of ${\bm{W}}_{id,0}^{}({i_{th}},:)({\bm{W}}_{cam,0}^{}({j_{th}},:))$ is 0, the value at the corresponding position of ${\bm{m}}_{s,i}^{id}({\bm{m}}_{s,i}^c)$ is also 0. If the element at a certain position of ${\bm{W}}_{cam,0}^{}({j_{th}},:)({\bm{W}}_{id,0}^{}({i_{th}},:))$ is not 0, the element at the corresponding position of ${\bm{m}}_{s,i}^{id}({\bm{m}}_{s,i}^c)$ is 1. After the 0 in ${{\bm{\tilde f'}}_{s,i}}$ and ${{\bm{\bar f'}}_{s,i}}$ are deleted, the recombined features ${{\bm{\tilde f}}_{s,i}} \in \mathbb{R}^{1 \times n/2}$ and ${{\bm{\bar f}}_{s,i}} \in \mathbb{R}^{1 \times n/2}$ are obtained. At the same time, the 0 that appears in ${\bm W}_{id,0}^{}({i_{th}},:)$ is deleted due to deactivation, and the obtained pedestrian identity classifier is recorded as ${\bm{\tilde W}}_{id,0}^{}({i_{th}},:) \in \mathbb{R}^{1 \times n/2}$ . The same operation for camera classifier ${\bm{W}}_{cam,0}^{}$ is used to get the recombined camera classifier ${\bm{\tilde W}}_{cam,0}^{}({j_{th}},:) \in \mathbb{R}^{1 \times n/2}$.
\begin{figure}[t!]
\centering
\includegraphics[width=3.3in,height=1.8in]{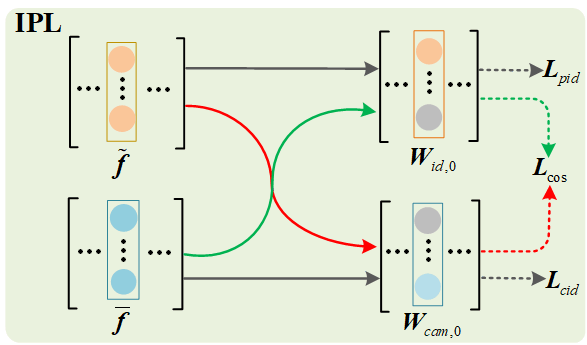}
\caption{Details of interactive promotion learning. The red and green solid lines respectively represent the pedestrian identity flow and the camera identity flow.}
\label{label4}
\end{figure}
\subsubsection{Interactive Promotion Module}
To further remove the style information of the camera from ${{\bm{\tilde f}}_{s,i}}$ and the identity information of pedestrians from ${{\bm{\bar f}}_{s,j}}$, the IPL mechanism is proposed. The details of the specific process are shown in Fig. \ref{label4}. Specifically, the recombined features ${{\bm{\tilde f}}_{s,i}}$ and ${{\bm{\bar f}}_{s,j}}$ are input to the corresponding recombined classifiers ${\bm{\tilde W}}_{id,0}^{}$ and ${\bm{\tilde W}}_{cam,0}^{}$ to update $\bm E$, so that the features extracted by $\bm E$ can still be classified to the corresponding pedestrian and camera identities after recombination.
\begin{equation}
\begin{aligned}
    \bm L_{id2}^{}({\bm{E}})
    &= \frac{1}{{{n_s}}}\sum\limits_{i = 1}^{{n_s}} Ce\Big({\bm{\tilde W}}_{id,0}^{}\big({{{\bm{\tilde f}}}_{s,i}}\big),{{{y}}_{s,i}}\Big)\\
    &+ \frac{1}{{{n_s}}}\sum\limits_{i = 1}^{{n_s}} Ce\Big({\bm{\tilde W}}_{cam,0}^{}\big({{{\bm{\bar f}}}_{s,i}}\big),{c_{s,i}}\Big)
\end{aligned},
\end{equation}

In this process, the features that only provide weak contributions are removed. In this case, to ensure that ${\bm{\tilde W}}_{id,0}^{}$ and ${\bm{\tilde W}}_{cam,0}^{}$ obtained after recombination using the pre-trained classifier can still correctly classify the recombined ${{\bm{\tilde f}}_{s,i}}$ and ${{\bm{\bar f}}_{s,j}}$, the discriminability of features ${{\bm{\tilde f}}_{s,i}}$ and ${{\bm{\bar f}}_{s,j}}$ must be improved accordingly. Therefore, minimizing Eq.(11) can improve the ability of $\bm E$ to extract discriminative features. In fact, for a specific task, the above method improves the generalization performance of the model by limiting the number of model parameters.

In the IPM, $\bm E$ is further updated so that the feature ${{\bm{\tilde f}}_{s,i}}$ extracted by $\bm E$ does not contain camera information, and ${{\bm{\bar f}}_{s,j}}$ does not contain pedestrian information. This objective can be achieved by minimizing the loss function as follows.
\begin{equation}
\begin{aligned}
    \bm L_{\cos }^{}({\bm{E}})
    &= \frac{1}{{{n_s}}}\sum\limits_{i = 1}^{{n_s}} Ce\Big({\bm{\tilde W}}_{id,0}^{}\big({{{\bm{\bar f}}}_{s,i}}\big),{\bar y_{s,i}}\Big) \\
    &+ \frac{1}{{{n_s}}}\sum\limits_{i = 1}^{{n_s}} Ce\Big({\bm{\tilde W}}_{cam}^{}\big({{{\bm{\tilde f}}}_{s,i}}\big),{\bar c_{s,i}}\Big)
\end{aligned},
\end{equation}
where ${\bar y_{s,i}} \in \mathbb{R}^{1 \times {N_s}}$, and the value of each element in ${\bar y_{s,i}}$ is $\frac{1}{{{N_s}}}$, ${\bar c_{s,i}} \in \mathbb{R}^{1 \times ({N_{{c_s}}} + {N_{{c_t}}})}$. The value of each element in ${\bar c_{s,i}}$ is $1/{N_{{c_s}}} + {N_{{c_t}}}$. The loss function used in the IPM can be expressed as follows.
\begin{equation}
\begin{aligned}
    \bm L_{pro}^{}({\bm{E}}) = {\bm L}_{id2}^{}({\bm{E}}) + {\bm L}_{\cos }^{}({\bm{E}})
\end{aligned},
\end{equation}
\subsection{Cross-camera Consistency Learning}
This paper aims to address the domain-adaptive person re-ID problem due to the lack of cross-camera paired samples. In this setting, there are no pedestrian across cameras with the same identity in the training samples from the target dataset due to the long distance between cameras. Therefore, it is challenging to learn discriminative features across cameras under the supervision of unpaired samples across cameras. This paper addresses the above problem from two aspects, instance-level distribution-consistency learning of source-domain and target-domain features and cross-camera identity consistency learning.
\subsubsection{Inter-domain Instance-level Distribution Consistency}
To make the target-domain samples acquired across long-distance scenes more effectively participate in the model training, an instance-level distribution consistency implementation method is proposed based on the style transfer theory of $AdaIN$ \cite{57}. Let ${{\bm{f}}_{s,i}}$ and ${{\bm{f}}_{t,j}}$ be the features of ${\bm x_{s,i}}$ and $\bm{x_{t,j}}$ extracted by $\bm E$. The corresponding mean values are $\mu ({{\bm{f}}_{s,i}})$ and $\mu ({{\bm{f}}_{t,j}})$, respectively. The corresponding variances are $\sigma ({{\bm{f}}_{s,i}})$ and $\sigma ({{\bm{f}}_{t,j}})$, respectively. Then, after the style is transferred from source domain to target domain, the features can be expressed as follows.
\begin{equation}
\begin{aligned}
    \bm f_{s,i}^{s \to t} = \sigma (\bm f_{t,j}^{})\frac{{\bm f_{s,i}^{} - \mu (\bm f_{s,i}^{})}}{{\sigma (\bm f_{s,i}^{})}} + \mu (\bm f_{t,j}^{})
\end{aligned},
\end{equation}
\begin{equation}
\begin{aligned}
    \bm f_{t,j}^{t \to s} = \sigma (\bm f_{s,i}^{})\frac{{\bm f_{t,j}^{} - \mu (\bm f_{t,j}^{})}}{{\sigma (\bm f_{t,j}^{})}} + \mu (\bm f_{s,i}^{})
\end{aligned},
\end{equation}

The inconsistency of feature distribution between source-domain data and target-domain data is the main factor causing domain shift. If the source-domain and target-domain sample features have the same distribution, it will be conducive to narrowing the domain shift between the source domain and the target domain. However, diverse factors cause style differences in samples. Eliminating the impact of domain discrepancy on a single sample is challenging, when distribution alignment is directly implemented at the domain level. Therefore, the instance-level distribution-consistency learning is presented. The specific process is given as follows.
\begin{equation}
\begin{aligned}
    \bm L_{Div}({{\bm{f}}_{t,j}},{\bm{f}}_{_{t,j}}^{t \to s}) = D_{KL}\Big(P\big({{\bm{f}}_{t,j}}\big)\left\| {Q} \right.\big({\bm{f}}_{_{t,j}}^{t \to s}\big)\Big)
\end{aligned},
\end{equation}
\begin{equation}
\begin{aligned}
	\bm L_{Div}({{\bm{f}}_{s,i}},{\bm{f}}_{s,i}^{s \to t}) = D_{KL}\Big(P\big({{\bm{f}}_{s,i}}\big)\left\| {Q} \right.\big({\bm{f}}_{s,i}^{s \to t}\big)\Big)
\end{aligned},
\end{equation}
\begin{equation}
\begin{aligned}
	\bm L_{dir\_style}(E) = \bm L_{Div}({{\bm{f}}_{t,j}},{\bm{f}}_{_{t,j}}^{t \to s}) + \bm L_{Div}({{\bm{f}}_{s,i}},{\bm{f}}_{s,i}^{s \to t})
\end{aligned},
\end{equation}
where $P$ and $Q$ are the probability distributions of the input features, and $D_{KL}( \cdot || \cdot )$ is the KL divergence, which is utilized to evaluate the difference between the two distributions. The inter-domain variance is reduced by minimizing Eq.(18) to narrow down the domain distribution between the source and target domains.

In the above process, since the sample style of the target domain is transferred to the source-domain samples, the positive samples across cameras in the target domain are generated, which allows the model to be trained supervised in the style of the target domain. Therefore, the following loss function is used to update $\bm E$.
\begin{equation}
\begin{aligned}
    \bm L_{id}^{trans}({\bm{E}}) = \frac{1}{{{n_s}}}\sum\limits_{i = 1}^{{n_s}} Ce\Big({{\bm{W}}_{id}}\big({\bm{f}}_{s,i}^{s \to t}\big),{y_{s,i}}\Big)
\end{aligned},
\end{equation}

In the above process, the total loss of obtaining inter-domain distribution consistency can be formulated as follows.
\begin{equation}
\begin{aligned}
	\bm L_{style}({\bm{E}})
    &=\bm L_{dir\_style}({\bm{E}}) + \bm L_{id}^{trans}({\bm{E}})\\
    &+ \frac{1}{{{n_t}}}\sum\limits_{j = 1}^{{n_t}} Ce\Big({\bm{W}}_{cam,0}^{}\big({\bm{f}}_{t,j}^{}\big),{\bar c}_{t,j}\Big) \\
    &+ \frac{1}{{{n_s}}}\sum\limits_{i = 1}^{{n_s}} Ce\Big({\bm{W}}_{cam,0}^{}\big({\bm{f}}_{s,i}^{}\big),{\bar c}_{s,j}\Big)
\end{aligned},
\end{equation}
\subsubsection{Identity Consistency across Cameras}
After the above process, the model has a certain domain-adaptive ability in the target domain, but it does not fully consider the difficult sample classification on the target dataset. Therefore, this paper proposes an effective method to solve how isolated camera samples in the target domain participate in model training. However, pedestrians with the same identity across cameras generally do not appear across long-distance scenes. This means that the samples of the same identity across cameras are unavailable to make the model learn identity-consistent features across cameras.

In fact, samples with the same identity across cameras can be regarded as similar ones. If the model can reduce the difference between similar samples, it can also make the distance between the same pedestrian sample features across cameras smaller.
Based on this idea, this paper proposes that the similarity between the same identity across cameras can be maximized by clustering the target-domain features and enhancing the similarity of samples within the same class.

Therefore, a cluster-based method for learning identity-consistent features across cameras is developed, aiming at solving the problem of missing positive samples across cameras. Specifically, ${\bm x_{t,j}}$ is symbolized as the $j$-th sample of the target domain, and the identity label ${y}_{t,j}$ of ${\bm x_{t,j}}$ is obtained after clustering. To enable $\bm E$ to extract more similar features from pedestrian images with the same identity acquired from different cameras, this paper proposes to minimize the following loss to update $\bm E$.
\begin{equation}
\begin{aligned} 
	\bm L_{cluster}^{}({\bm{E}},{{\bm{W}}_{t,id}})
	&= \frac{1}{{{n_t}}}\sum\limits_{j = 1}^{{n_t}} Ce\bigg({{\bm{W}}_{t,id}}\Big({\bm{{\bm E}}}\big(\bm {x_{t,j}}\big)\Big),{y_{t,j}}^{}\bigg)\\
	&+ \frac{1}{{{n_t}}}\sum\limits_{j = 1}^{{n_t}} Ce\Big({\bm{W}}_{cam,0}^{}\big({{\bm{f}}_{t,j}}\big),{\bar c}_{t,j}\Big)
\end{aligned},
\end{equation}
where  ${{\bm{W}}_{t,id}}$ is the identity classifier for samples on the target domain.

Minimizing Eq.(21) can make the features of similar samples show a higher similarity. However, hard negative samples with different identities are not considered. Since pedestrian tracking can be applied to quickly obtain the identity information of each tracked pedestrian appeared in a single camera \cite{58,59}, the identity of pedestrians within the same camera views is easy to know. Assume that  ${\tilde y}_{t,j}$ is the identity label of the sample ${\bm x_{t,j}}$ in a single camera view. The following loss function can be conducted to solve the problem that hard negative samples cannot be distinguished.
\begin{equation}
\begin{aligned}
	\bm L_{id3}^{}({\bm{E}},{{\bm{\tilde W}}_{t,id}}) = \frac{1}{{{n_t}}}\sum\limits_{j = 1}^{{n_t}} Ce\bigg({{{\bm{\tilde W}}}_{t,id}}\Big({\bm{E}}\big({\bm x_{t,j}}\big)\Big),{\tilde y}_{t,j}\bigg)
\end{aligned},
\end{equation}
\begin{equation}
\begin{aligned}
    \bm L_{tri}^{}(\bm E) =  - \frac{1}{{{n_t}}}\sum\limits_{j = 1}^{{n_t}} {\max (0,m + ||{{\bm{f}}_{t,j}}}  - {\bm{f}}_{t,j}^p|{|_2} - ||{{\bm{f}}_{t,j}} - {\bm{f}}_{t,j}^n|{|_2})
\end{aligned},
\end{equation}
where ${\bm{f}}_{t,j}^n$ and ${\bm{f}}_{t,j}^p$ are the hard negative sample features of ${\bm x_{t,j}}$ in a mini-batch and the hard positive sample features from the same camera respectively. $m$ is the threshold. ${{\bm{\tilde W}}_{t,id}}$ is the identity classifier of pedestrians on the target dataset.

In cross-camera identity feature consistency learning, the complete loss function is given as follows.
\begin{equation}
\begin{aligned}
    \bm L_{real}({\bm{E}}) = \bm L_{id3}^{}({\bm{E}},{{\bm{\tilde W}}_{t,id}}) +\bm  L_{tri}^{}({\bm{E}}) + \bm L_{cluster}^{}({\bm{E}},{{\bm{W}}_{t,id}})
\end{aligned}.
\end{equation}
\begin{algorithm}[!t]\small
\caption{\textbf{} Domain-adaptive Person Re-ID without Paired Samples across Cameras}\label{alg:A}
\begin{algorithmic}
\STATE {\textbf{Input:} Source-domain samples $\bm {D_s} = \{\bm x_{s,i}^{},y_{s,i}^{},c_{s,i}^{}\} \left| {_{i = 1}^{{N_s}}} \right.$, target-domain sample $\bm {D_t} = \{ \bm x_{t,j}^{},c_{t,j}^{}\} \left| {_{j = 1}^{{N_t}}} \right.$, pedestrian identity labels under an isolated camera: $\bm Y_{}^t = \left\{{y_{t,j}^{}} \right\}\left| {_{j = 1}^{{n_t}}} \right.$, pedestrian classifier $\bm{W}_{id}$, camera classifier ${{\bm{W}}_{cam}}$.\\}
\STATE {\textbf{Output:} The trained feature extractor $\bm E$.\\
\begin{flushleft}
\textbf{Step \uppercase\expandafter{\romannumeral1}:} Pre-training (Sec.\uppercase\expandafter{\romannumeral3}.B)\\
~1:Sample a batch of labeled source data.\\
~2:Sample a batch of unlabeled target data.\\
~3:Initialize $\bm E$, ${{\bm{W}}_{id}}$, ${{\bm{W}}_{cam}}$.\\
~4: \textbf{for} \emph{iter}=1, $\cdots$, \emph{Iteration}$_{1}$ \textbf{do}\\
~5:\qquad Update $\bm E$ and ${{\bm{W}}_{id}}$ by minimizing the loss in Eq.(1).\\
~6:\qquad Update ${{\bm{W}}_{cam}}$ by minimizing the loss in Eq.(2).\\
~7: \textbf{end for}\\
\textbf{Step \uppercase\expandafter{\romannumeral2}:} Category Synergy Co-promotion Module (Sec.\uppercase\expandafter{\romannumeral3}.C)\\
~8:Sample a batch of labelled source data.\\
~9:Load the learned $\bm E$, ${{\bm{W}}_{id}}$, ${{\bm{W}}_{cam}}$.\\
10: \textbf{for} \emph{iter}=1, $\cdots$, \emph{Iteration}$_{2}$ \textbf{do}\\
11:\qquad Update $\bm E$ by minimizing the loss in Eq.(13).\\
12: \textbf{end for}\\
\textbf{Step \uppercase\expandafter{\romannumeral3}:} Cross-camera Consistency Learning  (Sec.\uppercase\expandafter{\romannumeral3}.D)\\
13:Sample a batch of labelled source data.\\
14:Sample a batch of unlabeled target data.\\
15:Load the learned $\bm E$, ${{\bm{W}}_{id}}$, ${{\bm{W}}_{cam}}$.\\
16:Initialize the classifier ${{\bm{W}}_{t,id}}$ and ${{\bm{\tilde W}}_{t,id}}$.\\
17: \textbf{for} \emph{iter}=1, $\cdots$, \emph{Iteration}$_{3}$ \textbf{do}\\
18:\qquad Update $\bm E$ by minimizing the loss in Eq.(20).\\
19:\qquad Update $\bm E$ by minimizing the loss in Eq.(24).\\
20: \textbf{end for}\\
\end{flushleft}}
\end{algorithmic}
\end{algorithm}
\subsection{Algorithm and Optimization}
In the training phase, the samples from source domain are used in the first 100 epochs to perform model pre-training by minimizing Eq.(1). The camera classifier is pre-trained by minimizing Eq.(2) in the 100-200 epochs. Feature distribution alignment is achieved by minimizing Eq.(20) in the 200-300 epochs. In the remaining 100 epochs, the learning of consistent features of the same identity across cameras is achieved by minimizing Eq.(24). In the testing phase, $\bm E$ is used for feature extraction, and Euclidean distance is performed for pedestrian identity matching. The optimization algorithm is given in \textbf{Algo. 1}.

\begin{table*}[!ht]\small
\centering {\caption{Details of the datasets used in the following experiments.}\label{tabl111}
\renewcommand\arraystretch{1.2}
\begin{tabular}{|c|c|c|c|c|c|c|c|}
\hline
\multirow{1}{*}{Datasets} & \multirow{1}{*}{Cam} & \multirow{1}{*}{Training IDs} & \multirow{1}{*}{Training Imgs} & \multirow{1}{*}{Testing IDs} & \multirow{1}{*}{Testing Imgs} & \multirow{1}{*}{Easy Annotation} & \multirow{1}{*}{Paired data Across Cameras}\\
\hline
  Market-1501 &6&751&12,936&750&15,913 &$\times$
   &\checkmark \\
  \hline
  Market-SCT &6&751&3,561&750&15,913 &\checkmark &$\times$ \\
  \hline
  DukeMTMC	&8&702&16,522&1,110&17,661 &$\times$ &\checkmark \\
  \hline
  Duke-SCT	&8&702&5,993&1,110&17,661&\checkmark&$\times$ \\
  \hline
  MSMT17	&15&1,041&32,621&3,060&93,820&$\times$&\checkmark \\
  \hline
  MSMT-SCT	&15&1,041&6,694&3,060&93,820&\checkmark&$\times$ \\
  \hline
\end{tabular}}
\end{table*}
\begin{table*}[!ht]\small
\centering {\caption{Comparison of experiments on Market$\rightarrow$Duke-SCT and Duke$\rightarrow$MSMT-SCT after adding different modules. The mAP, Rank-1, Rank-5, and Rank-10 results are listed.}\label{tabl222}
\renewcommand\arraystretch{1.3}
\begin{tabular}{|c|c|c|c|c|c|c|c|c|}
\hline
\multirow{2}*{Methods} & \multicolumn{4}{c|}{Market$\rightarrow$Duke-SCT} & \multicolumn{4}{c|}{Duke$\rightarrow$MSMT-SCT} \\
\cline{2-9}
&Rank-1 &Rank-5  &Rank-10&mAP  &Rank-1    &Rank-5 &Rank-10 &mAP\\
\hline
Baseline &66.8&79.5&83.2&47.9&39.0&51.5&57.9&16.3 \\
Baseline+FRT &69.9&82.3&86.8&52.2&43.5&56.7&62.9&20.2 \\
Baseline+FRT+IPL &78.6&87.9&90.7&63.8&44.0&57.7&63.7&20.5 \\
Baseline+FRT+IPL+FDA
&79.4&88.6&91.1&64.1&48.1&62.0&68.0&24.6	\\
Baseline+FRT+IPL+FDA+ICL
&83.0&91.2&93.8&69.1&56.5&69.1&74.6&31.1	\\
  \hline
\end{tabular}}
\end{table*}
\section{Experiments}
\subsection{Datasets and Evaluation Protocol}
To verify the effectiveness of the proposed method, this paper uses three datasets specialized for person re-ID across distant scenes, namely Market-SCT \cite{1}, DukeMTMC-SCT (Duke-SCT) \cite{1} and MSMT17-SCT (MSMT-SCT) \cite{2} as the target datasets. These three datasets are derived by resetting the datasets Market-1501 (Market) \cite{27}, DukeMTMC (Duke) \cite{28}, and MSMT17 \cite{29} according to the characteristics of across distant scenes. In Market, Duke, and MSMT17, each pedestrian image in the training set has the corresponding samples with the same identity across cameras. However, in Market-SCT, Duke-SCT, and MSMT17-SCT, the training set only has single-camera pedestrian samples, and there are no samples with the same pedestrian identity across cameras. In person re-ID across long-distance scenes, the datasets Market, Duke, and MSMT17 are used as source-domain data, and Market-SCT, Duke-SCT, and MSMT17-SCT are used as target-domain data. The specific information of different datasets is shown in Table \ref{tabl111}. In the objective evaluation of algorithm performance, Cumulative Matching Characteristics (CMC) \cite{60} and mean Average Precision (mAP) \cite{27} are used to evaluate the performance of each method under a single query setting.
\subsection{Implementation Details}
This paper uses the TransReID \cite{56} framework as the backbone of the proposed method. All pedestrian images are resized to 256$\times$128 and the batch size is 16. During model training, operations such as horizontal flipping, padding, random cropping, and random erasing are used for data augmentation. The SGD optimizer \cite{61} with a momentum of 0.9, a weight decay rate of $1 \times {10^{ - 4}}$, and a learning rate of $1.6 \times {10^{ - 3}}$ are used to update the parameters of the encoder $\bm E$. In this process, the learning rate of ${{\bm{W}}_{id}}$, ${{\bm{W}}_{cam}}$, ${{\bm{W}}_{t,id}}$ and ${{\bm{\tilde W}}_{t,id}}$ is $1.2 \times {10^{ - 4}}$. The proposed method is implemented under the PyTorch framework \cite{62}, and all experiments are completed on the platform with one NVIDIA GeForce RTX 2080 Ti GPU. The model is trained for a total of 400 epochs. In the first 10 epochs, the learning rate is linearly adjusted by the warm-up strategy \cite{63}. At the 40-th and 70-th epochs, the learning rate decays by 10\%.
\begin{table*}[!ht]\small
\centering {\caption{The performance of the proposed method is compared with that of state-of-the-art methods on Market-SCT and Duke-SCT. The recognition accuracies on mAP, Rank-1, Rank-5, and Rank-10 are listed. The experimental comparisons of USL and UDA methods are sorted according to the year of publication. "--" indicates no reported data. The bold font indicates the optimal data, and the underline indicates the suboptimal data.}\label{tabl333}
\renewcommand\arraystretch{1.3}
\begin{tabular}{|c|c|c|c|c|c|c|c|c|c|c|}
\hline
\multirow{2}*{Methods} & \multicolumn{5}{c|}{Market-SCT} & \multicolumn{5}{c|}{Duke-SCT} \\
\cline{2-11}
&Source&Rank-1 &Rank-5&Rank-10&mAP&Source&Rank-1 &Rank-5&Rank-10&mAP\\
\hline
  \textbf{USL} &&&&&&&&&&\\
  MCNL(AAAI'20)\cite{1}	&None&67.0&82.8&87.9&41.6&None&67.1&80.9&84.7&45.2\\
  Precise-ICS(WACV'21)\cite{30} &None&50.0&67.5&74.8&31.2&None&41.2	&57.9&64.2&25.9\\
  AGW(TPAMI'21)\cite{32} &None&56.0&72.3&79.1&36.6&None&56.5	&71.0&77.7&43.9\\
  SimSiam(CVPR'21)\cite{33} &None&36.2&51.9&59.1&18.0&None	&28.1&43.2&51.3&19.7\\
  STS(arXiv'21)\cite{34} &None&21.1&34.3&41.3&8.5&None&33.0	&45.8&50.9&18.4\\
  ICE(ICCV'21)\cite{35} &None&29.3&41.1&47.2&13.4&None&20.4	&28.2&33.0&11.6\\
  CCFP(ACMMM'21)\cite{2} &None&82.4&92.6&95.4&63.9&None&\underline{80.3}	&\underline{89.0}&\underline{91.9}&\underline{64.5}\\
  CCSFG(CVPR'22)\cite{3} &None&\underline{84.0}&\underline{94.3}&\underline{96.2}&\underline{67.7}&None&--&--&--&--\\
  PPLR(CVPR'22)\cite{36} &None&17.9&26.1&30.9&6.0&None&15.8	&22.2&26.8&8.4\\
   \hline
  \textbf{UDA}                &&&&&&&&&&\\
  MMT-500(ICLR'20)\cite{18} &Duke&50.0&68.0&75.9&27.8&Market&38.9&56.3&63.5&26.8\\
  MMT-700(ICLR'20)\cite{18}
  &Duke&49.1&66.9&74.3&27.7&Market&40.9&58.1&65.5&29.2\\
  MMT-900(ICLR'20)\cite{18}
  &Duke&51.0&70.0&76.9&28.5&Market&42.3&59.6&67.6&30.4\\
  SPCL(NeurIPS'20)\cite{37}
  &Duke&11.5&23.5&30.2&4.5&Market&12.3&19.7&24.2&5.6\\
  Meb-Net(ECCV'20)\cite{38}
  &Duke&54.4&71.1&78.1&30.7&Market&41.6&58.1&64.0&27.8\\
  CAC(INS'21)\cite{39}
  &Duke&62.1&76.6&81.1&30.6&Market&49.6&64.0&69.8&30.0\\
  IDM(ICCV'21)\cite{40}
  &Duke&32.3&48.3&56.1&14.3&Market&37.9&51.2&58.4&23.6\\
  Dual-Refine(TIP'21)\cite{41}
  &Duke&47.7&63.4&70.1&23.3&Market&39.8&53.4&60.2&28.1\\
  P2LR(AAAI'22)\cite{42}
  &Duke&52.6&68.5&75.1&25.9&Market&35.7&49.8&56.4&20.6\\
  DRDL(KBS'22)\cite{17}
  &Duke&60.8&76.6&81.2&27.7&Market&63.4&75.1&78.3&41.6\\
  \textbf{Proposed}
  &Duke&\bf 86.3&\bf 94.4&\bf 96.7&\bf 68.3&Market&\bf 83.0&\bf 91.2&\bf 93.8&\bf 69.1\\
  \hline
\end{tabular}}
\end{table*}
\subsection{Ablation Study}
The proposed method consists of four core components, feature recombination by task (FRT) and interactive promotion learning (IPL), feature distribution alignment (FDA), and identity consistent learning across cameras with single-camera supervision (ICL). The effectiveness of the above core components is verified on the Market$\rightarrow$Duke-SCT and Duke$\rightarrow$MSMT-SCT. After removing the above four core components from the proposed method, the obtained model is used as the baseline method, and the loss function shown in Eq.(1) is used to train the baseline model. In the ablation experiment, the method after adding FRT to Baseline is denoted as Baseline+FRT, the method after adding IPL to Baseline+FRT is denoted as Baseline+FRT+IPL, the method after adding FDA to Baseline+FRT+IPL is denoted as Baseline+FRT+IPL+FDA, the method of the complete model is denoted as Baseline+FRT+IPL+FDA+ICL.

\textbf{Effectiveness of FRT}. To improve the discriminability of features, this paper groups the features on different channels by task and designs FRT. To verify the effectiveness of FRT, FRT is added to Baseline and the performance of Baseline+FRT is compared with that of Baseline. As shown in Table \ref{tabl222}, compared with Baseline, the recognition performance obtained by Baseline+FRT on both Market$\rightarrow$Duke-SCT and Duke$\rightarrow$MSMT-SCT is significantly improved. Specifically, after adding FRT, the mAP increases by 4.3\% and 3.9\% on the Market$\rightarrow$Duke-SCT and Duke$\rightarrow$MSMT-SCT, respectively. Additionally, the recognition accuracy on Rank-1 also improves from 66.8\% (39.0\%) to 69.9\% (43.5\%). This indicates that FRT plays a positive role in improving model performance.

\textbf{Effectiveness of IPL}. IPL is mainly used to remove camera-related information from features, thereby reducing the limitation of the interference information brought by the camera style on person re-ID. According to the experimental results shown in Table \ref{tabl222}, when IPL is added to Baseline+FRT, the corresponding performance of the model on the Market$\rightarrow$Duke-SCT (Duke$\rightarrow$MSMT-SCT) increases from 69.9\% and 52.2\% (43.5\% and 20.2\%) in Rank-1 and mAP to 78.6\% and 63.8\% (44.0\% and 20.5\%). This shows that IPL is effective in removing camera information to improve the domain adaptability of the model.

\textbf{Effectiveness of FDA}. To reduce the distribution inconsistency of source-domain and target-domain features caused by domain shift, FDA module is designed to achieve instance-level feature distribution alignment. As shown in Table \ref{tabl222}, after adding FDA to Baseline+FRT+IPL, the Rank-1 and mAP performance of Baseline+FRT+IPL+FDA on the Market$\rightarrow$Duke-SCT (Duke$\rightarrow$MSMT-SCT) changes from the original 78.6\% and 63.8\% (44.0\% and 20.5\%) to 79.4\% and 64.1\% (48.1\% and 24.6\%), respectively. This shows that the FDA plays a positive role in improving the domain adaptability of the re-ID model.
\begin{figure}[t!]
	\centering
	\includegraphics[width=3.0in,height=3.6in]{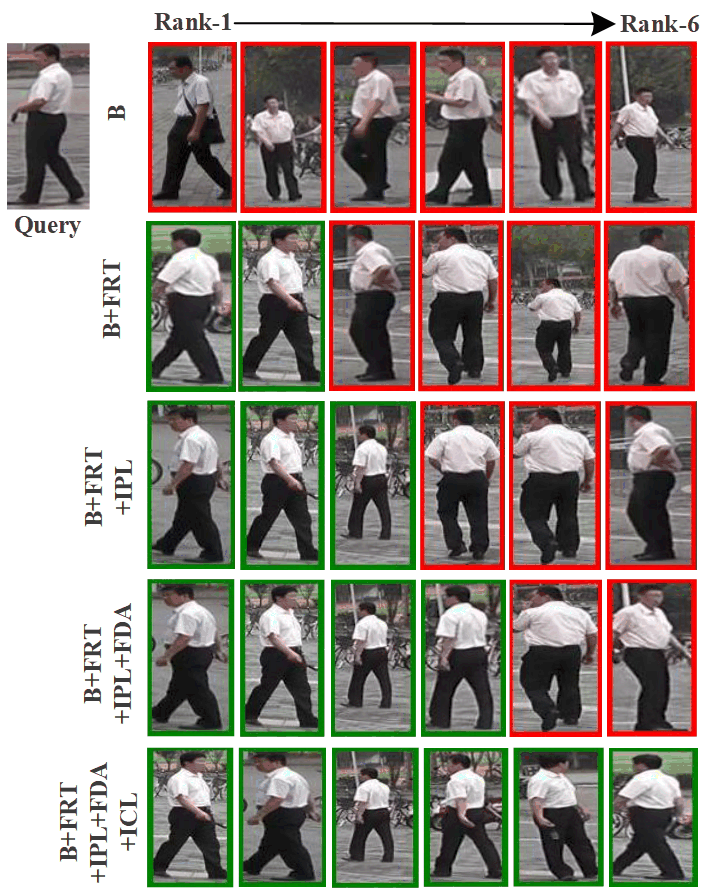}
	\caption{Visualization of retrieval results of Rank-1$\rightarrow$Rank-6 under different ablation settings on Market-SCT.}\label{figure5}
\end{figure}
\begin{figure}[t!]
	\centering
	\includegraphics[width=2.8in,height=2.1in]{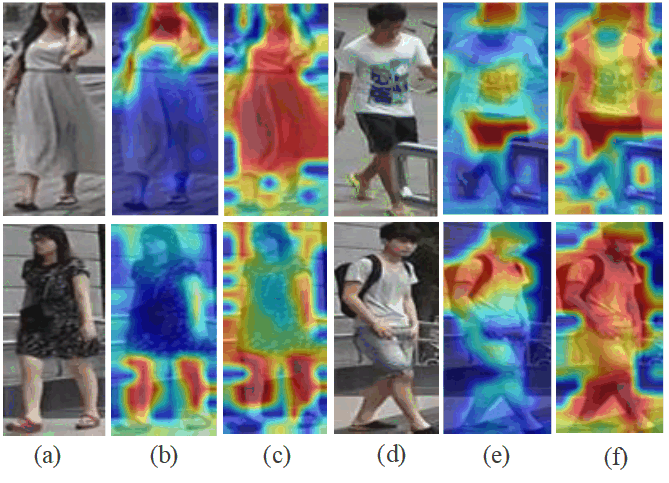}
	\caption{The impact of the proposed method. Columns (a) and (d) show the original pedestrian images. Columns (b) and (e) show the attention maps (heat maps) of $\textrm{Baseline}$. Columns (c) and (f) show the attention maps (heat maps) of $\textrm{Baseline+FRT+IPL+FDA+ICL}$. Warmer colors indicate stronger activation regions and colder colors indicate weaker activation regions.}
\label{figure6}
\end{figure}

\textbf{Effectiveness of ICL}. The ICL module is introduced to solve the problem  of missing cross-camera positive samples. As shown in Table \ref{tabl222}, compared with Baseline+FRT+IPL+FDA, the corresponding Rank-1 and mAP increase from 79.4\% and 64.1\% (48.1\% and 24.6\%) to 83.0\% and 69.1\% (56.5\% and 31.1\%) on Market$\rightarrow$Duke-SCT (Duke$\rightarrow$MSMT-SCT) respectively, after the introduction of ICL. The increase rates of Rank-1 and mAP reach 3.6\% and 5.0\% (8.4\% and 6.5\%), respectively. This is because ICL enables samples with similar features to show higher similarity. Additionally, it can effectively extend the distance between hard negative samples across cameras, which plays a positive role in pedestrian identity matching. Fig. \ref{figure5} shows the pedestrian retrieval results after using different modules, from which the contribution of each module is intuitively illustrated. Fig. \ref{figure6} also qualitatively shows the degree of contribution of the proposed method. Through quantitative and qualitative experiments, the effectiveness of each module in the proposed method has been demonstrated.
\subsection{Comparison with State-of-the-Art Methods}
\textbf{Experiments on Market-SCT and Duke-SCT}. To verify the effectiveness of the proposed method and its superiority over existing methods, Market-SCT and Duke-SCT are used as the target datasets, and Duke and Market are used as the source-domain datasets respectively. In these experiments, the comparison methods mainly involve both unsupervised learning (USL) and unsupervised domain adaptive (UDA) person re-ID methods. USL methods mainly include MCNL \cite{1}, Precise-ICS \cite{30}, AGW \cite{32}, SimSiam \cite{33}, STS \cite{34}, ICE \cite{35}, CCFP \cite{2}, CCSFG \cite{3}, and PPLR \cite{36}. UDA methods mainly include MMT-900 \cite{18}, SPCL \cite{37}, Meb-Net \cite{38}, CAC \cite{39}, IDM \cite{40}, Dual-Refine \cite{41}, P2LR \cite{42}, and DRDL \cite{17}. The results listed in Table \ref{tabl333} were obtained by using the codes provided by the corresponding original authors and retraining the corresponding models under the original parameter settings. All comparative experiments only used the optimal preset parameters provided by authors in their papers.

For the USL methods, CCSFG achieves the suboptimal performance on Market-SCT. Its Rank-1 and mAP reach 84.0\% and 67.7\%, respectively. CCFP also achieves the suboptimal performance on Duke-SCT, with Rank-1 and mAP reaching 80.3\% and 64.5\%, respectively. In contrast, Rank-1 and mAP obtained by the proposed method are 86.3\% and 68.3\% (83.0\% and 69.1\%) respectively on the Market-SCT (Duke-SCT). Compared with the second-best performance, the proposed method improves performance by 2.3\% and 0.6\% (2.7\% and 4.6\%) on Market-SCT (Duke-SCT), respectively. This shows that the proposed method has better recognition performance in person re-ID across long-distance scenes than the above-mentioned USL methods. The proposed method belongs to the domain-adaptive recognition methods. Compared with the domain-adaptive methods listed in the Table \ref{tabl333}, the proposed method also shows better performance. This is mainly because the existing methods are affected by the lack of paired training samples of the target data, which limits their performance on Market-SCT and Duke-SCT.
\begin{table*}[!ht]\small
	\centering {\caption{The performance of the proposed method and the existing USL methods on MSMT-SCT is compared. The bold in the table indicates the optimal data, and the underline indicates the suboptimal data.}\label{tabl444}
		\renewcommand\arraystretch{1.3}
		\begin{tabular}{|c|c|c|c|c|c|}
			\hline
			\multirow{2}*{Methods} & \multicolumn{5}{c|}{MSMT-SCT} \\
			\cline{2-6}
			&Source&Rank-1 &Rank-5&Rank-10&mAP \\
			\hline
			\textbf{USL}  &&&&&\\
			MCNL(AAAI'20)\cite{1}	&None &26.6 &40.0 &46.4	&10.0 \\
			Precise-ICS(WACV'21)\cite{30} &None &17.2 &28.4 &34.3	&6.7 \\
			AGW(TPAMI'21)\cite{32} &None &23.0 &33.9 &40.0	&11.1 \\
			SimSiam(CVPR'21)\cite{33} &None &2.8 &5.9	&8.4 &1.2 \\
			STS(arXiv'21)\cite{34} &None &13.2	&21.3 &25.9	&4.7 \\
			ICE(ICCV'21)\cite{35} &None &10.6	&17.6 &21.5	&4.0 \\
			CCFP(ACMMM'21)\cite{2} &None &50.1	&63.3 &68.8	&22.2\\
			CCSFG(CVPR'22)\cite{3} &None &\underline{54.6} &\underline{67.7} &\underline{73.1}	&\underline{24.6} \\
			PPLR(CVPR'22)\cite{36} &None &5.3	&9.5 &12.4 &2.1\\
			\textbf{Proposed} &Duke &\bf 56.5&\bf 69.1 &\bf 74.6 &\bf 31.1 \\
			\hline
	\end{tabular}}
\end{table*}
\begin{table*}[!ht]\small
	\centering {\caption{The performance comparison between the proposed method and the existing UDA methods on MSMT-SCT. The bold in the table indicates the optimal data, and the underline indicates the suboptimal data.}\label{tabl555}
		\renewcommand\arraystretch{1.3}
		\begin{tabular}{|c|c|c|c|c|c|c|c|c|}
			\hline
			\multirow{2}*{Methods} & \multicolumn{4}{c|}{Duke$\rightarrow$MSMT-SCT} & \multicolumn{4}{c|}{Market$\rightarrow$MSMT-SCT} \\
			\cline{2-9}
			&Rank-1 &Rank-5&Rank-10&mAP&Rank-1 &Rank-5&Rank-10&mAP\\
			\hline
			\textbf{UDA} &&&&&&&&\\
			MMT-1000(ICLR'20)\cite{18} &16.1&26.1&31.5&6.5&15.8&26.0&31.5&6.4\\
			MMT-2000(ICLR'20)\cite{18} &15.9&25.5&30.7&6.3&16.8&26.5&32.1&6.7\\
			SPCL(NeurIPS'20)\cite{37} &8.4 &14.5&18.5&3.2&5.4&10.5&13.3&2.0 \\
			Meb-Net(ECCV'20)\cite{38} &21.7&32.6&37.9	&7.4&14.5&23.4&28.5&4.9 \\
			CAC(INS'21)\cite{39} &35.6&48.0&53.7&12.6&29.0&40.6&46.3&10.6\\
			IDM(ICCV'21)\cite{40} &11.8&18.7&23.1	&4.6&11.3&18.7&22.9	&4.6\\
			Dual-Refine(TIP'21)\cite{41} &18.7&29.1&34.2	&6.7&12.9&21.3&26.1&4.9\\
			P2LR(AAAI'22)\cite{42} &23.6&34.3&39.9	&8.0&15.9&24.6&29.5&5.4\\
			DRDL(KBS'22)\cite{17} &\underline{41.9}&\underline{54.3}&\underline{60.0}&\underline{15.6}&\underline{36.0}&\underline{48.0}&\underline{53.4}&\underline{13.6}\\
			\textbf{Proposed} &\bf 56.5&\bf 69.1&\bf 74.6&\bf 31.1&\bf 51.4&\bf 65.3&\bf 71.2&\bf 28.3\\
			\hline
	\end{tabular}}
\end{table*}

\textbf{Experiments on MSMT-SCT}. To further prove the effectiveness of the proposed method, MSMT-SCT is used as the target dataset, and both Duke and Market are used as the source-domain datasets to train the model in the second set of experiments. Since MSMT-SCT contains more samples than Market and Duke, Duke$\rightarrow$MSMT-SCT and Market$\rightarrow$MSMT-SCT are more challenging. To verify the superiority of the proposed method over the USL methods, the proposed method is first compared with the USL-based methods, including MCNL \cite{1}, Precise-ICS \cite{30}, AGW \cite{32}, SimSiam \cite{33}, STS \cite{34}, ICE \cite{35}, CCFP \cite{2}, CCSFG \cite{3}, and PPLR \cite{36}, and the experimental results of different methods are shown in Table \ref{tabl444}. It can be seen that the proposed method outperforms state-of-the-art USL methods. In addition, compared with the domain-adaptive methods MMT-1000 \cite{18}, MMT-2000 \cite{18}, SPCL \cite{37}, Meb-Net \cite{38}, CAC \cite{39}, IDM \cite{40}, Dual-Refine \cite{41}, P2LR \cite{42}, and DRDL \cite{17}, the proposed method also shows strong competitiveness. As shown in Table \ref{tabl555}, Rank-1 and mAP obtained by the proposed method reach 56.5\% and 31.1\% (51.4\% and 28.3\%) on Duke$\rightarrow$MSMT-SCT (Market$\rightarrow$MSMT-SCT) respectively. Compared with the second-best performance, the proposed method improves performance by 14.6\% and 15.5\% (15.4\% and 14.7\%), respectively. This shows that the proposed method has more advantages on the task of spanning from small-size datasets to large-size datasets.
 \begin{figure}[!t] \centering
	\subfigure[]  {\includegraphics[height=1.5in,width=1.6in,angle=0]{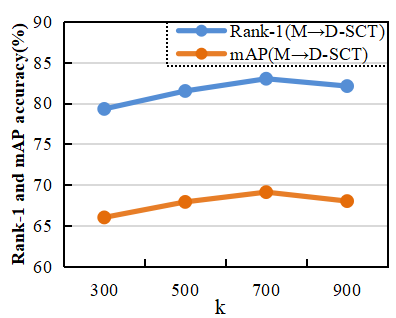}}
	\subfigure[]   {\includegraphics[height=1.5in,width=1.6in,angle=0]{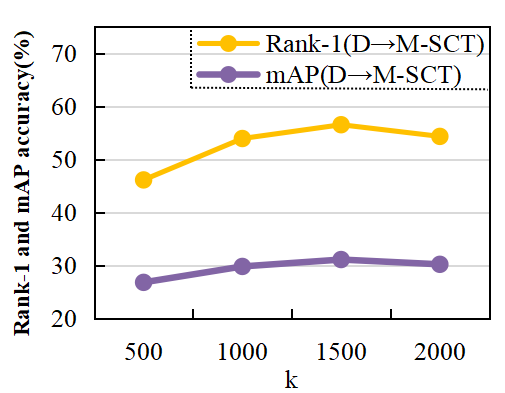}}
	\caption{Performance analysis with different values of hyperparameters. (a) On Market$\rightarrow$Duke-SCT, Rank-1 and mAP accuracy when the number of clusters k takes different values. (b) On Duke$\rightarrow$MSMT-SCT, Rank-1 and mAP accuracy when the number of clusters k takes different values.}\label{figure7}
\end{figure}

\subsection{Parameter Selection and Analysis}
In the proposed method, clustering is adopted to make the features between similar samples have high similarity, so that the model is able to emphasize feature similarity between camera views. Since the optimal solution of hyperparameters cannot be obtained through an optimization method, the optimal hyperparameter values are searched through experiments in this study. Fig. \ref{figure7} shows the changes in model performance when k takes different values. In the clustering process, the performance of the proposed model is affected, when the number of clusters is too large or too small. When k=700 (k=1,500), the proposed method achieves the best performance on Market$\rightarrow$Duke-SCT (Duke$\rightarrow$MSMT-SCT). Therefore, in the experiments conducted on Market$\rightarrow$Duke-SCT, k is set to 700, while in the experiments conducted on Duke$\rightarrow$MSMT-SCT, k is set to 1,500.

\section{Conclusion}
Aiming at the challenges of person re-ID across long-distance scenes, this paper proposes a domain-adaptive person re-ID method without cross-camera paired samples. This method addresses the above challenges by pedestrian discriminative feature learning, source-domain and target-domain distribution consistency learning, and cross-camera consistency learning. In pedestrian discriminative feature learning, this paper proposes a feature-by-task recombination mechanism. In accordance with the contribution of features to pedestrian identification and camera identification, this mechanism realizes the recombination of features according to tasks, and further model training strengthens the discriminability of recombined features on the corresponding tasks. Furthermore, to achieve the exclusion of feature information from other tasks in specific task-specific features, an interactive promotion learning mechanism is proposed to facilitate feature refinement. Additionally, based on the style transfer theory, this paper achieves the distribution alignment of source-domain and target-domain features. This paper also designs a cross-camera feature consistency learning mechanism to solve the challenge brought by the absence of cross-camera paired samples. A large number of experimental results confirm the effectiveness of the proposed method, and the ablation experiments show the effectiveness of each core component of the proposed model.
\bibliography{mybibfile1}
\end{document}